\documentclass[letterpaper]{article} 
\usepackage[preprint]{aaai2027}  
\usepackage[hyphens]{url}  
\usepackage{graphicx} 
\usepackage{natbib}  
\usepackage{caption} 
\usepackage{algorithm}
\usepackage{algorithmic}
\usepackage{mathtools}
\usepackage{amssymb}
\usepackage[table]{xcolor}
\usepackage{colortbl}
\usepackage{newfloat}
\usepackage{listings}
\DeclareCaptionStyle{ruled}{labelfont=normalfont,labelsep=colon,strut=off} 
\floatstyle{ruled}
\newfloat{listing}{tb}{lst}{}
\floatname{listing}{Listing}

\usepackage{booktabs}

\title{Stability-Aligned Residual Adaptation for Rapid Recovery from Dynamics Shifts}

\author{
Nethmi Jayasinghe\textsuperscript{\rm 1},
Diana Gontero\textsuperscript{\rm 1},
Francesco Paolo Migliarba\textsuperscript{\rm 1},\\
Spencer T.~Brown\textsuperscript{\rm 2},
Vinod K.~Sangwan\textsuperscript{\rm 3},
Mark C.~Hersam\textsuperscript{\rm 3,4,5},\\
Amit Ranjan Trivedi\textsuperscript{\rm 1}
}
\affiliations{}

\usepackage{multirow}
\begin{document}

\maketitle

\begingroup
\renewcommand{\thefootnote}{}
\footnotetext{%
\raggedright
\footnotesize
\textsuperscript{1}Department of Electrical and Computer Engineering,
University of Illinois Chicago, Chicago, IL, USA.

\textsuperscript{2}Department of Neurobiology,
Northwestern University, Evanston, IL, USA.

\textsuperscript{3}Department of Materials Science and Engineering,
Northwestern University, Evanston, IL, USA.

\textsuperscript{4}Department of Electrical and Computer Engineering,
Northwestern University, Evanston, IL, USA.

\textsuperscript{5}Department of Chemistry,
Northwestern University, Evanston, IL, USA.

Correspondence: Nethmi Jayasinghe (wjayas3@uic.edu),
Amit Ranjan Trivedi (amitrt@uic.edu).
}
\endgroup

\begin{abstract}
Robotic systems deployed beyond controlled environments encounter abrupt and unobserved dynamics shifts that induce substantial transient performance loss, even when the nominal closed loop remains locally stabilizing. We formulate rapid inference-time recovery for physical AI as a constrained residual learning problem, in which an online correction added to a frozen reinforcement learning policy is regulated to minimize recovery time while remaining inside the robustness margin of the nominal controller. We specifically focus upon the regulation of residual's corrective authority relative to an already competent frozen policy. The learned residual produces bounded action-space corrections online, and a Stability Alignment Gate regulates this corrective authority through magnitude constraints, directional coherence, performance-conditioned regulation, and adaptive gain modulation. The method requires no environment resets, policy retraining, explicit system identification, or privileged shift information. Across mid-episode changes in actuation, mass, and contact conditions, it reduces recovery time relative to frozen SAC by up to \textbf{87\%} on Go1, \textbf{48\%} on Cassie, \textbf{30\%} on H1, and \textbf{20\%} on Scout in simulation, while preserving near-nominal steady-state performance. The framework is also characterized on two \emph{physical} robots, a Trossen ViperX manipulator and an AgileX Scout Mini, cutting recovery time by \textbf{19.5\%} on hardware without modifying onboard controller or accessing shift information. 
\end{abstract}

\section{Introduction}
\label{sec:intro}

Learned control policies for physical AI systems often fail when deployed
outside their training conditions. Changes in actuation, mass, or contact
properties can cause abrupt performance degradation, requiring recovery
during execution without resetting the system or retraining the policy.
Such recovery is challenging when the deployed policy is frozen for safety or
certification, since modifying its parameters or latent conditioning can
alter the closed-loop behavior that provides nominal competence. We therefore
study recovery around a frozen policy: restoring task performance after a dynamics shift of unknown identity and magnitude without modifying the nominal.

Runtime adaptation methods typically modify either the policy or add an
external correction. Meta-learning and test-time adaptation update policy
parameters or inferred context online
\cite{finn2017model,rakelly2019efficient,wang2021tent}, while classical
adaptive control updates controller gains
\cite{ioannou1996robust,Slotine:1991:ANC}. These approaches couple recovery
with changes to the deployed controller. Residual learning preserves the
policy by adding corrective actions
\cite{johannink2019residual}, but an unconstrained online residual can oppose
stabilizing actions and degrade recovery.

This observation motivates the central question of this work. Once a
competent policy is frozen and a residual channel is available, the key
challenge is regulating how much \emph{corrective authority} the residual
receives and which directions it is allowed to influence. A fixed
high-dimensional feature basis already provides sufficient capacity to
represent useful corrections. We hypothesize that online adaptation is
primarily limited by corrective-authority regulation rather than residual
capacity. We introduce the Stability Alignment Gate (SAG), which regulates residual
authority through four mechanisms: global and per-joint magnitude bounds,
directional coherence with the nominal action, performance-conditioned global
gain, and per-joint amplification based on sustained error. The residual is
generated using a fixed feature expansion with dual-timescale online updates.
Its separation of fixed representation and error-driven correction is
inspired by cerebellar motor organization; this analogy is provided only as
architectural motivation and detailed in the appendix. SAG preserves the
nominal policy mapping while allowing bounded residual correction to reshape
post-shift recovery.

We evaluate the framework across four simulated morphologies (quadruped,
biped, humanoid, and wheeled platforms) and two physical robots, a Trossen
ViperX manipulator and an AgileX Scout Mini. The method accelerates recovery
without modifying the deployed controller or accessing dynamics-shift
information. This paper contributes:
(i) a formulation of inference-time recovery as constrained residual
adaptation where corrective-authority regulation is the central variable;
(ii) the Stability Alignment Gate, a mechanism that regulates residual
influence and analytically reduces interference with nominal actions; and
(iii) an evaluation across diverse robot platforms, including a factorial
analysis separating the effects of corrective authority and residual
capacity.

\section{Relation to Prior Work}

Prior work on runtime adaptation either improves robustness during training or
modifies the deployed controller during execution. We instead study
bounded residual adaptation around a frozen policy.

\vspace{2pt}
\noindent\textbf{Robust RL.}
Domain randomization and adversarial training
\cite{tobin2017domain,pinto2017robust,peng2018sim} expose policies to
parameter variations during training, but the deployed controller remains
fixed and does not adapt post-shift.

\vspace{2pt}
\noindent\textbf{Online Identification and Meta-Adaptation.}
RMA, OSI, MAML, PEARL, and test-time training
\cite{kumar2021rma,yu2017preparing,qiu2025umcunifiedresilientcontroller,finn2017model,rakelly2019efficient}
adapt online by inferring latent context or updating policy parameters,
thereby changing the deployed controller or its conditioning.

\vspace{2pt}
\noindent\textbf{Adaptive Control and Robust MPC.}
MRAC, RLS, and disturbance observers
\cite{ioannou1996robust,Slotine:1991:ANC} update controller gains or estimate
system parameters under assumptions such as known model structure and
persistent excitation, which are difficult to establish for learned policies.
Robust MPC incorporates uncertainty models directly into the controller.

\vspace{2pt}
\noindent\textbf{Residual and Cerebellar-Inspired Architectures.}
Residual policy learning adds corrective actions to a nominal controller
\cite{johannink2019residual}, while cerebellar-inspired models introduce
error-driven corrections over baseline programs
\cite{Albus1971,KawatoGomi1992,albus1975new,jayasinghe2026cerebellarinspiredresidualcontrolfault}.
These approaches provide corrective mechanisms, whereas we focus on
regulating deployment-time influence of an online residual on a
frozen policy.

\begin{figure*}[!t]
  \centering
  \includegraphics[width=2.1\columnwidth]{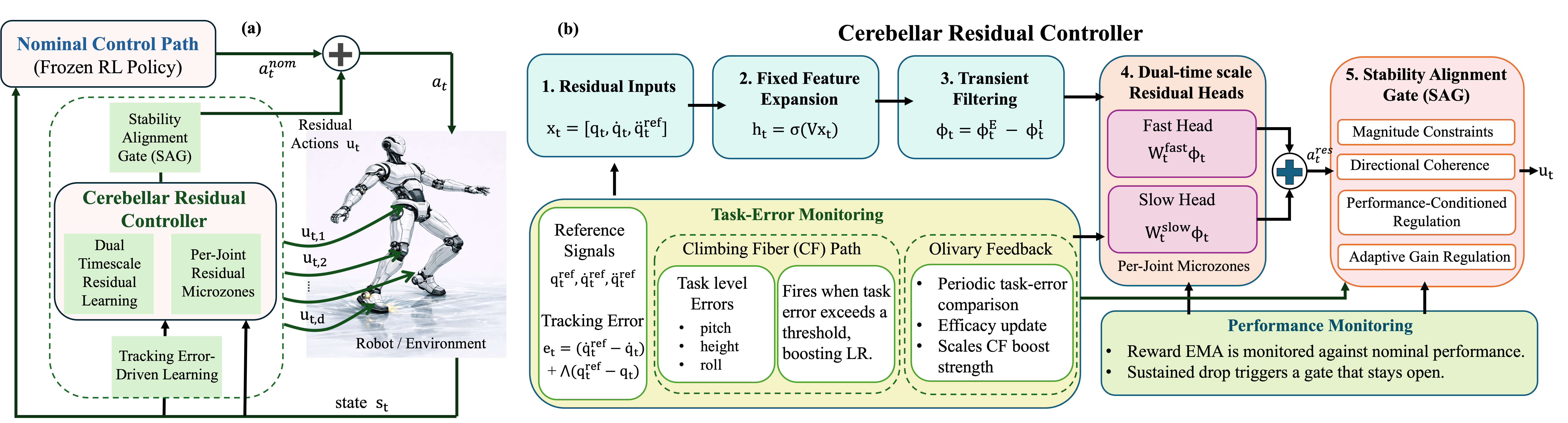}
  \vspace{-5pt}
\caption{Overview of authority-regulated residual adaptation. (a) A frozen RL policy generates nominal actions, while a parallel residual channel learns online corrections from tracking errors; the Stability Alignment Gate (SAG) regulates their injection before forming the final action. (b) Residual inputs pass through a fixed feature expansion, transient filtering, and fast and slow per-joint heads. Task-error and performance signals modulate the online residual update and the corrective authority, while SAG enforces magnitude bounds, directional coherence, performance-conditioned regulation, and adaptive gain regulation.\vspace{-10pt}}

  \label{fig:architecture}
\end{figure*}

\section{Authority-Regulated Residual Adaptation}
\label{sec:method}

We present the framework in three parts: the constrained recovery problem
(\ref{sec:problem_def}), the residual learner that generates candidate
corrections (\ref{sec:residual_channel}), and the Stability Alignment Gate
(SAG) that regulates their corrective authority (\ref{sec:sag}).

\subsection{Problem Formulation and Control Objective}
\label{sec:problem_def}

Consider a robotic system with state $s_t \in \mathcal{S}$ and control input
$a_t \in \mathcal{A}$ evolving under parameterized dynamics
\begin{equation}
s_{t+1}=f(s_t,a_t;\omega_t),
\end{equation}
where $\omega_t \in \Omega$ represents physical parameters such as actuator
gains, mass distribution, or contact properties. During training,
$\omega_t=\omega^\star$, and a policy $\pi_\theta$ is optimized under the
nominal closed loop
\begin{equation}
s_{t+1}=f(s_t,\pi_\theta(s_t);\omega^\star).
\end{equation}

\vspace{2pt}
\noindent\textbf{Assumption 1 (Local Robust ISS).}
There exists a neighborhood $\mathcal{W}\subset\Omega$ containing
$\omega^\star$ such that, for all $\omega\in\mathcal{W}$, the closed loop
induced by $\pi_\theta$ is locally input-to-state stable (ISS) \cite{sontag1989smooth} around an
invariant manifold $\mathcal{M}$.

\vspace{2pt}
\noindent\textbf{Justification.}
Learned locomotion policies are typically trained to stabilize trajectories
around a gait manifold $\mathcal{M}$ under bounded disturbances and modeling
noise. We therefore consider perturbations within a local robustness region
$\mathcal{W}$ where trajectories remain bounded despite transient performance
degradation. This assumption represents an empirical operating regime rather
than a certified property of the neural policy and does not imply global
robustness.

At deployment, an abrupt dynamics shift of unknown identity and magnitude
occurs at time $\tau$, with $\omega_t=\omega^\star$ for $t<\tau$ and
$\omega_t=\tilde{\omega}\neq\omega^\star$ for $t\geq\tau$. The goal is to
restore performance without modifying the nominal policy. The deployed
controller therefore takes the form
\begin{equation}
a_t=\pi_\theta(s_t)+u_t,
\end{equation}
where $u_t$ is generated online during execution.

Let $\bar{J}_t$ denote a smoothed performance signal and $J^\star$ the nominal
performance level. For tolerance $\delta>0$, define the recovery time
\begin{equation}
T_{\delta}\coloneqq
\inf\left\{t\geq\tau\mid \bar{J}_t\geq J^\star-\delta\right\}.
\end{equation}

The objective is to design an online residual correction $u_t$ that minimizes
$T_\delta$ under unknown post-shift dynamics $\tilde{\omega}$ while
preserving the nominal controller. We constrain the residual magnitude,
\begin{equation}
\|u_t\|_2\leq\varepsilon,\qquad |u_{t,j}|\leq\varepsilon_j,
\end{equation}
so that adaptation remains within a bounded correction envelope. The resulting
constrained residual learning problem is
\begin{equation}
\min_{\{u_t\}_{t\geq\tau}} T_{\delta}
\quad\text{s.t.}\quad
\|u_t\|_2\leq\varepsilon,\;
|u_{t,j}|\leq\varepsilon_j,\;
u_t=g_\psi(s_{\leq t}),
\label{eq:constrained_problem}
\end{equation}
where $g_\psi$ generates the residual online from observable signals without
access to $\tilde{\omega}$. Since a fixed feature basis can already provide
sufficient residual capacity, the central challenge becomes regulating the
magnitude and direction of corrective influence so that adaptation accelerates
recovery while preserving nominal behavior.

\subsection{Residual Channel and Authority Regulation}
\label{sec:residual_channel}

The design operates on the following principle: preserve a competent stabilizing controller and
introduce adaptation through a separate, authority-limited correction channel.
Prior test-time updates~\cite{sun2020ttt,wang2021tent},
meta-learned adaptation~\cite{finn2017model,rakelly2019efficient}, and
classical adaptive control~\cite{Slotine:1991:ANC,ioannou1996robust} modify
the deployed controller, which can perturb the nominal closed loop or require
system identification assumptions. Residual control provides an alternative
by adding corrective actions around a nominal controller
\cite{johannink2019residual}. We use the same separation principle and focus
on regulating the deployment-time influence of an online residual.

The learned policy $\pi_\theta$ is the nominal stabilizing controller,
while adaptation occurs through a parallel residual:
\begin{align}
a_t &= a_t^{\mathrm{nom}} + u_t,
\qquad
a_t^{\mathrm{nom}} \coloneqq \pi_\theta(s_t), \\
u_t &= \gamma_t \big(\beta_t \odot a_t^{\mathrm{res}}\big),
\label{eq:residual_injection}
\end{align}
where $a_t^{\mathrm{res}}\in\mathbb{R}^d$ is the residual output,
$\gamma_t\in\mathbb{R}_+$ controls global corrective authority, and
$\beta_t\in\mathbb{R}_+^d$ provides per-joint amplification. The residual
does not modify the nominal policy and instead provides bounded additive
correction:
\begin{equation}
\|u_t\|_2 \leq \varepsilon,
\qquad
|u_{t,j}| \leq \varepsilon_j .
\label{eq:residual_bound}
\end{equation}
These constraints keep adaptation within a bounded correction envelope around
the nominal closed loop.

\vspace{4pt}
\noindent\textbf{Residual Learner.}
The residual is generated from a fixed high-dimensional feature
expansion $h_t=\sigma(Vx_t)$ with fixed $V$, making adaptation an
error-driven reweighting of a static basis. A platform-specific feature
$\phi_t$ feeds fast and slow heads:
\[
a_t^{\mathrm{res}}
=
W_t^{\mathrm{fast}}\phi_t
+
W_t^{\mathrm{slow}}\phi_t .
\]
Distinct learning-rate scales and decay let the fast head capture
transients while the slow head retains persistent corrections. The
heads are updated online using platform-specific tracking or task error
with $\sigma$-modification leakage~\cite{ioannou1996robust}. The
reference requires a joint-space, velocity, or task-space target.
Detailed rules are provided in Supplementary Sec.~B.

\subsection{Stability Alignment and Authority Regulation}
\label{sec:sag}

An unconstrained residual can interfere with the nominal controller
$\pi_\theta$, making recovery slower or unstable. SAG regulates the
corrective authority of the residual through four coupled mechanisms:
magnitude constraints from Eq.~\eqref{eq:residual_bound}, directional
coherence with the nominal action, performance-conditioned global
regulation through $\bar{J}_t$, and adaptive gain modulation through
$\gamma_t$ and $\beta_t$. These mechanisms increase corrective
influence during degradation while preserving bounded intervention around
the nominal controller.

\vspace{4pt}
\noindent\textbf{Directional Coherence.}
Residual components that oppose the nominal action can cancel stabilizing
control effort. Since the nominal policy remains locally stabilizing within
the robustness region $\mathcal{W}$, SAG measures alignment independently in
each action dimension:
\begin{equation}
c_{t,j}
=
\operatorname{sgn}\!\left(a_{t,j}^{\mathrm{nom}}\right)
\operatorname{sgn}\!\left(a_{t,j}^{\mathrm{res}}\right),
\qquad j=1,\ldots,d .
\label{eq:joint_alignment}
\end{equation}
A negative value indicates an opposing residual component. SAG attenuates
only these components:
\begin{equation}
\rho(c_{t,j}) =
\begin{cases}
1, & c_{t,j}\geq0,\\
\kappa, & c_{t,j}<0,
\end{cases}
\qquad
u_{t,j}\leftarrow \rho(c_{t,j})u_{t,j},
\label{eq:joint_attenuation}
\end{equation}
where $\kappa\in[0,1)$ controls the retained opposing correction. Soft
attenuation preserves necessary corrective actions when a dynamics shift
requires partial opposition to the nominal command, while reducing
unnecessary interference during recovery. The \emph{No Dir. Align} ablation
in Table~\ref{tab:Ab_results} corresponds to $\kappa=1$ and evaluates the
effect of removing this constraint.

\begin{table*}[!t]                        
\centering
\small
\setlength{\tabcolsep}{2pt}    
\renewcommand{\arraystretch}{1.1} 

\begin{tabular}{llcccccccccccc}
\toprule
\textbf{Group} & \textbf{Method}
& \multicolumn{4}{c}{\textbf{Actuator Degradation (0.76$\times$)}}
& \multicolumn{4}{c}{\textbf{Mass Increase (1.15$\times$)}}
& \multicolumn{4}{c}{\textbf{Friction Increase (2.1$\times$)}} \\
\cmidrule(lr){3-6} \cmidrule(lr){7-10} \cmidrule(lr){11-14}
& & AUC$\uparrow$ & SSR$\uparrow$ & TTR-50$\downarrow$ & TotalR$\uparrow$
  & AUC$\uparrow$ & SSR$\uparrow$ & TTR-50$\downarrow$ & TotalR$\uparrow$
  & AUC$\uparrow$ & SSR$\uparrow$ & TTR-50$\downarrow$ & TotalR$\uparrow$ \\
\midrule

\shortstack{\textbf{No}\\\textbf{Adapt}}
& SAC-Only
& 0.437 & 0.427 & $\geq 4500$ & 9562$\pm$122
& 1.001 & 1.001 & 1950 & 18411$\pm$104
& 0.934 & 0.933 & $\geq 4500$ & 17233$\pm$115 \\
\midrule

\rowcolor{gray!15}
\multirow{4}{*}{\textbf{Online}}
& \textbf{SAC + Ours}
& 0.664 & 0.666 & 3962 & 13520$\pm$150
& \textbf{1.042} & \textbf{1.044} & \textbf{168} & 20150$\pm$295
& \textbf{0.973} & 0.965 & \textbf{189} & 18783$\pm$143 \\

& SAC+MRAC
& 0.372 & 0.392 & 2442 & 8439$\pm$3719
& 0.908 & 0.904 & 2929 & 17778$\pm$1498
& 0.496 & 0.445 & 4361 & 10595$\pm$1807 \\

& SAC+RLS
& 0.402 & 0.462 & \textbf{2230} & 8961$\pm$3443
& 0.918 & 0.912 & 2553 & 17957$\pm$1619
& 0.529 & 0.443 & 4102 & 11164$\pm$3447 \\

& Online SAC
& 0.384 & 0.243 & 3802 & 8651$\pm$3511
& 0.716 & 0.558 & 3813 & 14432$\pm$3159
& 0.432 & 0.325 & 4356 & 9470$\pm$2450 \\
\midrule

\multirow{6}{*}{\shortstack{\textbf{Spec.}\\\textbf{Train}}}
& OSI
& 0.555 & 0.554 & $\geq 4500$& 13458$\pm$1410
& 0.991 & 0.987 & 3556 & \textbf{22289$\pm$145}
& 0.972 & \textbf{0.967} & $\geq 4500$ & \textbf{21895$\pm$144} \\

& RMA
& 0.642 & 0.639 & $\geq 4500$ & 15208$\pm$2954
& 0.975 & 0.972 & $\geq 4500$ & 21923$\pm$1131
& 0.892 & 0.886 & $\geq 4500$ & 20246$\pm$1031 \\

& DR-SAC
& 0.571 & 0.384 & 3778 & 9578$\pm$3168
& 0.902 & 0.903 & 3334 & 14204$\pm$118
& 0.429 & 0.286 & 3727 & 7562$\pm$1481 \\

& PEARL
& 0.907 & 0.908 & 3512 & \textbf{18260$\pm$73}
& 0.964 & 0.964 & 2886 & 19282$\pm$71
& 0.869 & 0.869 & $\geq 4500$ & 17593$\pm$70 \\

& RARL
& \textbf{1.056} & \textbf{1.075} & 2260 & 10425$\pm$2388
& 0.511 & 0.472 & 2867 & 6086$\pm$4267
& 0.656 & 0.650 & 2711 & 6889$\pm$2618 \\

& RPL
& 0.716 & 0.717 & 3590 & 13591$\pm$170
& 1.014 & 1.013 & 1834 & 18490$\pm$133
& 0.894 & 0.894 & 2290 & 16519$\pm$131 \\
\bottomrule
\end{tabular}
\vspace{-5pt}
\caption{Go1 recovery performance under representative perturbations. Metrics include recovery AUC, SSR, TTR-50, and total return (mean $\pm$ std over 30 trials). Higher is better for AUC, SSR, and total return; lower is better for TTR-50.}\vspace{-15pt}
\label{tab:Go1_results}
\end{table*}


\vspace{4pt}
\noindent\textbf{Gain Regulation Dynamics.}
The global authority parameter $\gamma_t$ controls the overall influence of
the residual channel. It increases under performance degradation and
contracts as recovery occurs:
\begin{equation}
\gamma_t =
\mathrm{clip}_{[0,\gamma_{\max}]}
\left(
\gamma_{\min} +
k_\gamma
\frac{J^\star-\bar J_t}
{J^\star-J_{\min}+\xi}
\right).
\end{equation}
Here $J_{\min}$ is the minimum observed performance after the current
degradation onset and provides normalization relative to the severity of the
current shift. The $J_{\min}$ state is updated causally after adaptation is
activated and reset between rollouts.

Per-joint amplification further adjusts corrective authority:
\begin{equation}
\beta_{t+1,j} =
\mathrm{clip}_{[\beta_{\min},\beta_{\max}]}
\left(
\beta_{t,j}+
k_\beta\mathbb{I}\{|e_{t,j}|>\bar e_j\}
\right).
\end{equation}

Joints with persistent tracking error receive increased corrective emphasis.
The hysteresis band prevents rapid switching near the threshold. Both $\gamma_t$ and $\beta_t$ are clipped, and the final projection
enforces the residual bounds.

\vspace{2pt}
\noindent\emph{Proposition 1 (informal; full statement and proof in Supplementary Sec.~B.2).}
For $\kappa\in[0,1]$, the gated residual $\tilde u$ leaves the cooperative
component unchanged, $A(\tilde u)=A(u)$, contracts the antagonistic
component, $D(\tilde u)=\kappa D(u)$, and is nonexpansive,
$\|\tilde u\|_2\le\|u\|_2$. Hence the gate never increases interference with
the nominal action, reduces it whenever interference is present, and
cannot violate the magnitude bound.

\vspace{2pt}
\noindent\emph{Proof sketch.} The gate scales exactly the dimensions where
$a^{\mathrm{nom}}_j u_j<0$ by $\kappa\le 1$ and leaves the rest unchanged;
summing the per-dimension contributions gives the two identities, and
$\rho\in\{1,\kappa\}\subseteq[0,1]$ gives nonexpansiveness. A conditional first-order corollary in Supplementary
Sec.~B.2 relates $D(u)$ to manifold deviation.

We note this as an interpretive result about interference in action space,
not a stability theorem: it concerns agreement with the nominal \emph{action}
and presumes that action remains broadly stabilizing after the shift. It
isolates why setting $\kappa=1$ admits the largest antagonistic interference,
consistent with the ablation. Thus, SAG regulates residual injection through four coupled
mechanisms: bounded residual authority, directional coherence,
performance-conditioned regulation, and adaptive global and per-joint
gain modulation.
    
\begin{table*}[!tb]
\centering
\vspace{5pt}
\small
\setlength{\tabcolsep}{1.0pt}   
\renewcommand{\arraystretch}{1.1} 
\begin{tabular}{llcccccccccccc}
\toprule

& & \multicolumn{6}{c}{\textbf{Agility Cassie}} 
& \multicolumn{6}{c}{\textbf{Unitree H1}} \\

\cmidrule(lr){3-8} \cmidrule(lr){9-14}

& 
& \multicolumn{3}{c}{Mass (1.15$\times$)} 
& \multicolumn{3}{c}{Friction (1.4$\times$)}
& \multicolumn{3}{c}{Mass (1.15$\times$)} 
& \multicolumn{3}{c}{Friction (1.3$\times$)} \\

\cmidrule(lr){3-5} \cmidrule(lr){6-8}
\cmidrule(lr){9-11} \cmidrule(lr){12-14}

\textbf{Group} & \textbf{Method}
& SSR$\uparrow$ & TTR-50$\downarrow$ & TotalR$\uparrow$
& SSR$\uparrow$ & TTR-50$\downarrow$ & TotalR$\uparrow$
& SSR$\uparrow$ & TTR-50$\downarrow$ & TotalR$\uparrow$
& SSR$\uparrow$ & TTR-50$\downarrow$ & TotalR$\uparrow$ \\

\midrule

\shortstack{\textbf{No}\\\textbf{Adapt.}}
& SAC-Only 
& 1.037 & 1351 & 16755$\pm$2042 
& 0.958 & 2800 & 15568$\pm$458
& 1.064 & 298 & 24548$\pm$123 
& 1.060 & 465 & 24396$\pm$109 \\

\midrule

\rowcolor{gray!15}
\multirow{3}{*}{\textbf{Online}}
& \textbf{SAC + Ours} 
& \textbf{1.077} & 601 & 17219$\pm$1484
& \textbf{1.171} & \textbf{142} & 18234$\pm$442
& \textbf{1.065} & \textbf{116} & 25546$\pm$124 
& 1.061 & \textbf{373} & 25453$\pm$112 \\

& SAC+MRAC 
& 1.073 & 611 & 17151$\pm$844 
& 1.071 & 957 & 17477$\pm$746
& 1.062 & 392 & 25512$\pm$134
& \textbf{1.096} & 427 & 25784$\pm$245 \\

& SAC+RLS 
& 1.061 & 1051 & 16968$\pm$1101 
& 1.101 & 354 & 17749$\pm$732
& 1.063 & 505 & 25419$\pm$124 
& 1.069 & 436 & \textbf{25842$\pm$177} \\

\midrule

\multirow{3}{*}{\shortstack{\textbf{Spec.
}\\\textbf{Train}}}

& OSI 
& 0.844 & 2162 & 16147$\pm$3709
& 0.812 & 2822 & 17070$\pm$2083
& 0.931 & 2967 & 23128$\pm$1099 
& 1.039 & 1966 & 23495$\pm$731 \\

& RMA 
& 0.986 & 3900 & 16912$\pm$4419 
& 0.951 & 2942 & \textbf{18870$\pm$2873}
& 1.057 & 350 & 24573$\pm$233 
& 1.057 & 379 & 23633$\pm$400 \\

& DR-SAC 
& 1.070 & \textbf{594} & \textbf{17268$\pm$127} 
& 1.082 & 151 & 17573$\pm$119
& 1.060 & 334 & 23509$\pm$1545 
& 1.041 & 2043 & 23165$\pm$209 \\

\bottomrule
\end{tabular}
\caption{Cross-platform recovery performance on Agility Cassie (biped) and Unitree H1 (humanoid). Metrics include time-to-recovery at 50\% (TTR-50), steady-state ratio (SSR), and total episode return (TotalR $\pm$ std over 30 trials). Lower is better for TTR-50; higher is better for SSR and TotalR. \vspace{-10pt}} \label{tab:cross_platform_results}
\end{table*}

\section{Experimental Setup}

\noindent\textbf{Simulation Platform:}
Experiments are conducted in MuJoCo~\cite{todorov2012mujoco}. The primary
platform is a 12-DOF Unitree Go1 quadruped trained for forward locomotion
with normalized joint-position targets. Observations include joint states,
base-motion features, projected gravity, and previous actions.

\vspace{1pt}
\noindent\textbf{Nominal Policy:}
The nominal controller is a Soft Actor-Critic (SAC)~\cite{haarnoja2018sac}
policy trained under nominal dynamics $\omega^\star$ and frozen at
deployment. Evaluation permits no policy retraining, system identification,
reset-based recovery, or access to shift identity, magnitude, or physical
parameters.

\vspace{1pt}
\noindent\textbf{Mid-Episode Perturbation Protocol:}
Each rollout contains 5000 simulation steps, with a dynamics shift injected
at step 500 and maintained thereafter. We evaluate actuator scaling, mass
variation, and friction changes across six severity levels each (18 total
conditions), with 30 trials per condition.

\vspace{1pt}
\noindent\textbf{Evaluation Metrics:}
We report \textit{TTR-50}, the steps required for smoothed reward to recover
50\% of the post-shift drop (censored at 4500 steps), \textit{Recovery AUC},
\textit{Steady-State Ratio (SSR)}, and total episode reward. AUC and SSR are
normalized by the nominal pre-shift reward; values above 1.0 indicate improved
performance under the perturbed condition (e.g., reduced slip under higher
friction). TTR-50 uses identical smoothing across methods.

\vspace{1pt}
\noindent\textbf{Baselines:}
We compare against frozen SAC, online SAC updates, classical adaptive control
(MRAC, RLS), and robustness-through-training methods including DR-SAC, RMA,
OSI, PEARL, RARL, and RPL. All methods follow the same mid-episode
perturbation protocol without reset-based recovery. Supplementary Table~A2 summarizes information access,
training budgets, and deployment-time adaptation across
methods. Our method uses online
performance signals and tracking references but no shift identity, magnitude,
or physical parameters, and is trained only under nominal dynamics.
Baseline configurations are fixed per platform; training
budgets and selection protocols are detailed in
Supplementary Sec.~D. Methods
trained with dynamics variation (DR-SAC, RMA, OSI, PEARL, RARL) are frozen at
deployment; their training ranges and the corresponding in/out-of-distribution
conditions are in the appendix.

\begin{figure*}[!tb]
    \centering
    \includegraphics[width=\textwidth]{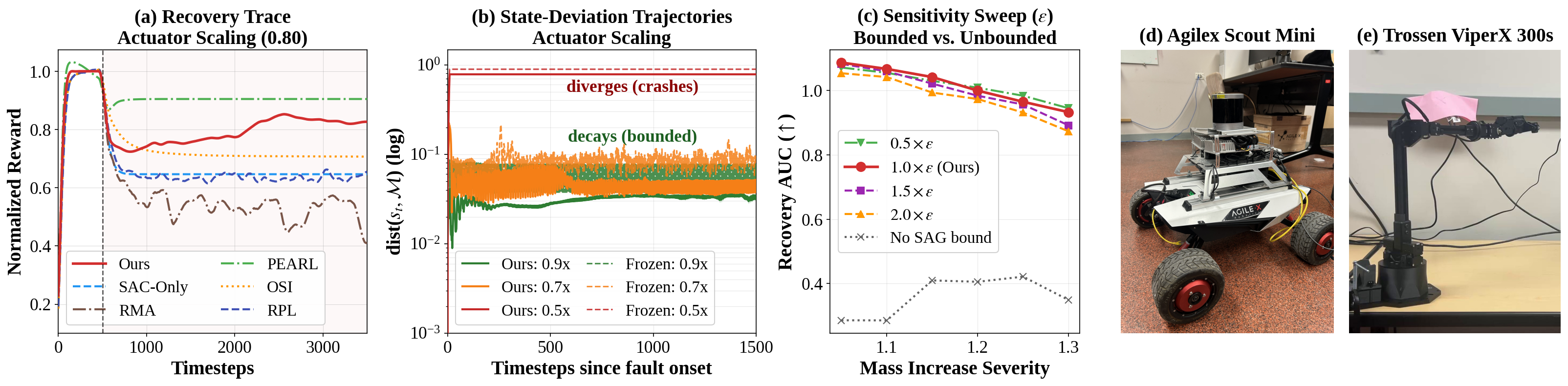}
    \vspace{-15pt}
    \caption{Go1 recovery, empirical boundedness, residual-authority
sensitivity, and hardware platforms. (a) Normalized reward after a
mid-episode $0.80\times$ actuator-scaling shift. (b) Deviation from
the nominal manifold under mild, boundary, and out-of-envelope shifts;
in-envelope trajectories remain bounded, while the most severe case
fails. (c) Recovery AUC across mass-increase severities and residual
bounds; removing the SAG bound degrades performance. Panels (a--c)
show Go1 results; (d,e) show the AgileX Scout Mini and Trossen
ViperX 300S used for hardware validation. \vspace{-15pt}
     } 
    \label{fig:recovery_trace}
\end{figure*}

\section{Experimental Evaluation}

We evaluate four aspects of the framework: (i) whether corrective-authority
regulation improves recovery and outperforms increasing residual capacity;
(ii) transfer across robot morphologies and dynamics shifts; (iii) physical
deployment; and (iv) the contribution of SAG components.

\subsection{Effect of Corrective Authority on Recovery}

Table~\ref{tab:Go1_results} compares recovery under representative actuator,
mass, and friction shifts. The proposed controller achieves the fastest recovery under
mass and friction shifts while improving actuator-shift AUC
and SSR.
Under actuator degradation ($0.76\times$), it achieves higher
AUC and SSR than the online adaptation baselines, although MRAC,
RLS, and Online-SAC reach TTR-50 sooner. Under a $1.15\times$ mass increase, TTR-50 decreases to
168 steps, while most baselines require thousands of steps or fail to recover
within the episode. Under a $2.1\times$ friction increase, the proposed method
again achieves the fastest recovery with TTR-50 of 189 steps. Under mass and friction shifts, near- or above-nominal SSR shows
that faster recovery does not reduce final performance. Relative to frozen SAC, SAG reduces TTR-50 by 91\% under
the mass shift; under friction, it reaches TTR-50 in 189
steps, whereas frozen SAC does not reach the threshold
within the recovery horizon. The
robustness-through-training baselines were exposed to dynamics variation
during training yet recover more slowly at deployment, indicating that
post-shift recovery speed is governed by how correction is applied online
rather than by prior exposure to the shift distribution. These results suggest that recovery quality depends on how the residual is
regulated rather than only on the residual learner's ability to represent
corrections. We next isolate corrective authority from residual capacity with
a factorial analysis.

\begin{table*}[!tb]
  \centering             
  \small                                 
  \begin{tabular}{lccccccc}
  \toprule
  & Full
  & No CF
  & No Dual-Head
  & No Olivary
  & No Persistence Gate
  & No Dir.\ Align
  & No Temp.\ Filt. \\
  \midrule
  TTR-50 $\downarrow$
  & \textbf{168}
  & 171
  & 186
  & 273
  & 174
  & 3367
  & 1127 \\

  AUC $\uparrow$
  & \textbf{1.042}
  & 1.040
  & 1.039
  & 1.041
  & 1.003
  & 1.021
  & 1.023 \\

  SSR $\uparrow$
  & \textbf{1.044}
  & 1.042
  & 1.005
  & 1.040
  & 1.002
  & 1.018
  & 1.024 \\
  
  TotalR $\uparrow$
  & \textbf{20150$\pm$295}
  & 20006$\pm$297
  & 19971$\pm$276
  & 20140$\pm$294
  & 20143$\pm$294
  & 19762$\pm$526
  & 19793$\pm$136 \\

  \bottomrule
  \end{tabular}
  \vspace{-5pt}
  \caption{Go1 ablation under a $1.15\times$ mass increase
($30$ trials). Lower TTR-50 and higher AUC, SSR, and TotalR are better.} 
  \label{tab:Ab_results}
  \end{table*}

\begin{table}[!tb]
\centering
\small
\setlength{\tabcolsep}{3.5pt}
\renewcommand{\arraystretch}{1.08}

\begin{tabular}{@{}lccc@{}}
\toprule
&
\shortstack{\textbf{ViperX}\\\textbf{Act. Dead zone}}
&
\shortstack{\textbf{ViperX}\\\textbf{Act. Scaling}}
&
\shortstack{\textbf{Scout Mini}\\\textbf{Emul. load}}
\\
\midrule

TTR-50 Nom. $\downarrow$
& $556 \pm 299$
& $498\pm244$
& 1000
\\

\rowcolor{gray!15}
TTR-50 Ours $\downarrow$
& \textbf{528} $\pm 292$
& \textbf{401} $\pm 186$
& \textbf{86}
\\

SSR Nom. $\uparrow$
& \textbf{0.869} $\pm 0.043$
& $0.857 \pm 0.055$
& $0.970$
\\

\rowcolor{gray!15}
SSR Ours $\uparrow$
& $0.863 \pm 0.045$
& \textbf{0.869} $\pm 0.044$
& \textbf{1.001}
\\
\bottomrule
\end{tabular}
\vspace{-5pt}
\caption{Physical-robot recovery under representative shifts.
ViperX reports mean $\pm$ standard deviation over $n=10$ trials;
Scout Mini reports one proof-of-concept trial. \vspace{-15pt}}
\label{tab:hardware}
\end{table}

\noindent\textbf{Corrective Authority versus Residual Capacity.}
\label{sec:factorial}
This experiment isolates corrective authority from residual capacity by
varying two axes independently. The first is an authority ladder that adds one
regulation mechanism at a time to a bare residual: $\mathrm{A}_0$ is the raw
residual; $\mathrm{A}_1$ adds the magnitude bound of
Eq.~\eqref{eq:residual_bound}; $\mathrm{A}_2$ adds directional coherence;
$\mathrm{A}_3$ adds performance-conditioned $\gamma_t$; and $\mathrm{A}_4$
adds per-joint $\beta_t$, giving the full gate. The second is a capacity axis
that varies only the fixed-feature width:
$\mathrm{C}_-$ uses $p=1000$, $\mathrm{C}_0$ uses the default $p=2500$, and
$\mathrm{C}_+$ uses $p=10000$. All cells use the same frozen SAC policy and
matched initializations, perturbations, and seeds, isolating authority and
capacity from policy training. Table~\ref{tab:factorial} reports TTR-50 on Go1
under a $2.1\times$ friction increase. Increasing capacity neither rescues the
ungated residual nor consistently improves the full gate, indicating that
recovery depends more on regulating residual influence than on increasing
residual width.

\subsection{Transfer Across Dynamics and Morphologies}
\begin{figure*}[!tb]
    \centering
    \includegraphics[width=\textwidth]{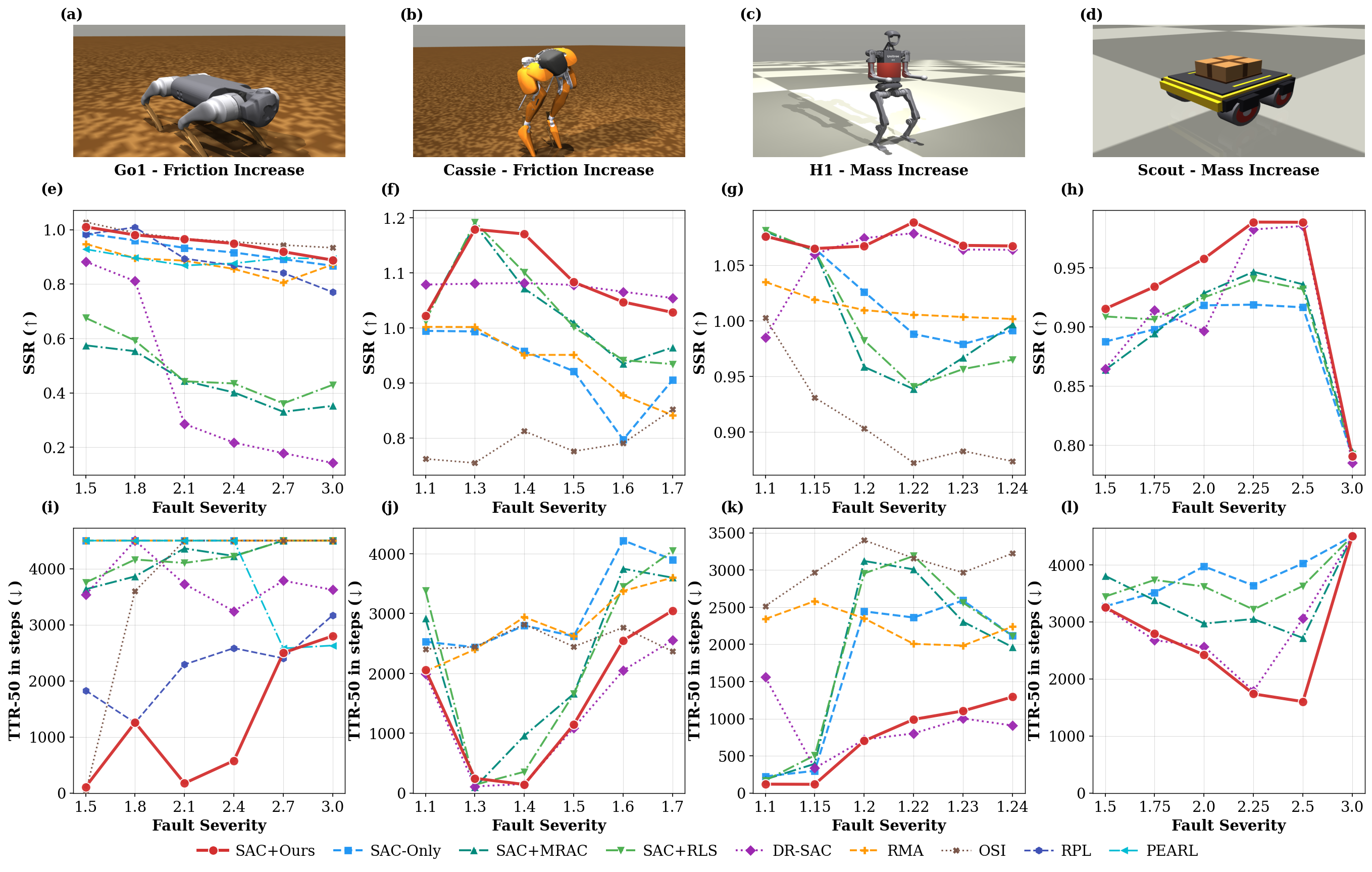}
\caption{Cross-platform recovery across increasing dynamics-shift
severity. (a--d) Representative perturbations for Go1, Cassie, H1,
and Scout. (e--h) Steady-state ratio (SSR; higher is better).
(i--l) TTR-50 (lower is better). Results are averaged over 30 trials
per condition. The proposed method generally accelerates recovery
while preserving near-nominal steady-state performance across robot
morphologies and control interfaces. \vspace{-10pt}}
    \label{fig:combined_fig}
\end{figure*}

\noindent\textbf{Severity scaling.} Figure~\ref{fig:recovery_trace}(a) shows that our controller produces
the steepest post-shift recovery and converges smoothly, whereas
several baselines exhibit oscillation or incomplete recovery. Across the
four-platform severity sweeps (Fig.~\ref{fig:combined_fig};
Go1 and Cassie friction increases, H1 and Scout mass increases),
the proposed method generally maintains competitive SSR while
reducing TTR-50 as severity grows. Performance degrades gradually over most
of the evaluated range, with the largest deterioration occurring near
the most severe perturbation limits. Complete severity sweeps are provided in Supplementary
Sec.~F.

Figure~\ref{fig:recovery_trace}(b) further compares state-deviation
trajectories for the residual-augmented and frozen controllers. Both
remain bounded under mild and near-boundary actuator shifts and fail
under the out-of-envelope shift, indicating that the residual improves
within-region settling without extending the nominal failure boundary.
Figure~\ref{fig:recovery_trace}(c) shows that recovery is relatively
insensitive to the precise bounded residual authority, whereas removing
the SAG bound substantially degrades performance.

\noindent\textbf{Cross-platform transfer.} We evaluate transfer across platforms with distinct
dimensionality, contact dynamics, and control interfaces.
Representative mass and friction conditions for Cassie and H1 are
reported in Table~\ref{tab:cross_platform_results}; the full severity sweeps for Cassie, H1, and Scout appear in
Fig.~\ref{fig:combined_fig}.

\smallskip
\noindent\textbf{Agility Cassie.} Under a $1.15\times$ mass increase, our method reduces TTR-50 from
1351 to 601 steps while improving SSR from 1.037 to 1.077. Under a
$1.4\times$ friction increase, TTR-50 decreases from 2800 to 142
steps and SSR improves from 0.958 to 1.171. Across the selected
friction sweep, the largest gains occur at moderate severities, with
performance degrading only near the largest shifts.

\smallskip
\noindent\textbf{Unitree H1.} On the 19-DOF humanoid, recovery improves while steady-state
performance remains near nominal. TTR-50 decreases from 298 to 116
steps under mass increase and from 465 to 373 steps under friction
increase. Across the selected mass sweep, SSR remains near nominal
while recovery time grows gradually with shift severity.

\smallskip
\noindent\textbf{AgileX Scout.} On the velocity-controlled wheeled platform, our method reduces
recovery time across moderate mass increases while maintaining
comparable or improved SSR. The advantage narrows at the most severe
shift, where available control authority becomes limiting and all
methods approach the recovery horizon.

Across quadruped, biped, humanoid, and wheeled platforms,
the same residual architecture transfers without shift-specific
models, while platform-level authority and adaptation
parameters are calibrated to each control interface.

\vspace{2pt}
\noindent\textbf{Compound Dynamics Shifts.}\label{sec:compound}
The shifts above are single-parameter. To test simultaneous combinations, we
evaluate two compound-shift families on Go1 at mild and severe settings:
mass--friction and actuator--friction. Both factors are injected mid-episode
without retuning, with $30$ trials per condition. Supplementary Table~F2
reports SSR and crash rate for the full method and frozen SAC. Under mild
combinations, the residual preserves or improves SSR without increasing crash
rate. Under severe shifts, both methods approach the failure regime, although
the residual improves SSR and reduces crashes for actuator--friction. These
results show that the residual can improve performance under compound shifts
without a shift-specific model, while remaining limited near the empirical
failure boundary.

\begin{table}[!t]
\centering
\small
\setlength{\tabcolsep}{3.2pt}
\renewcommand{\arraystretch}{1.08}
\begin{tabular}{@{}lccc@{}}
\toprule
\textbf{Authority} $\backslash$ \textbf{Capacity}
& $\mathbf{C}_{-}$
& $\mathbf{C}_{0}$
& $\mathbf{C}_{+}$ \\
& $p{=}1000$
& $p{=}2500$
& $p{=}10000$ \\
\midrule
Frozen SAC
& \multicolumn{3}{c}{$\geq 4500$} \\
\midrule
$\mathrm{A}_{0}$: Raw
& $\geq 4500$
& $\geq 4500$
& $\geq 4500$ \\

$\mathrm{A}_{1}$: $+$ Bound
& $3751{\pm}485$
& $3834{\pm}403$
& $777{\pm}388$ \\

$\mathrm{A}_{2}$: $+$ Direction
& $451{\pm}282$
& $448{\pm}138$
& $489{\pm}173$ \\

$\mathrm{A}_{3}$: $+$ Perf. reg.
& $400{\pm}202$
& $409{\pm}397$
& $448{\pm}183$ \\

\textbf{$\mathrm{A}_{4}$: Full SAG}
& $201{\pm}65$
& $\mathbf{189{\pm}90}$
& $339{\pm}282$ \\
\bottomrule
\end{tabular}
\vspace{-5pt}
\caption{Authority--capacity factorial on Go1 under
$2.1\times$ friction ($30$ trials). TTR-50 is mean
$\pm$ standard deviation in steps ($\downarrow$);
$\geq4500$ denotes no threshold crossing within the horizon.
Columns vary fixed-feature width $p$; rows add SAG mechanisms.\vspace{-15pt}}
\label{tab:factorial}
\end{table}

\subsection{Real-Robot Validations}

The statistical hardware evaluation is on a position-controlled Trossen
ViperX manipulator; the AgileX Scout Mini serves only as a controller-agnostic
feasibility check. In both cases the residual adapts online without modifying
the underlying controller or accessing shift identity or physical
parameters.

\vspace{2pt}
\noindent\textbf{Trossen ViperX.} We test actuator-scaling and actuator-dead-zone perturbations at three
severity levels, with 10 trials per condition. Table~\ref{tab:hardware}
reports representative results; the complete hardware sweep is provided in Supplementary
Table~G1. Our
method reduces recovery time by up to 19.5\% while preserving comparable
steady-state performance. Gains diminish under the most severe conditions
because the resulting loss of effective position-control authority limits
compensation.

\noindent\textbf{AgileX Scout Mini.} As a feasibility check, a single trial using a command-level emulation of
increased inertial load shows the residual reducing recovery time on the
wheeled platform. We report this only as qualitative evidence that the
approach is controller-agnostic, not as a hardware benchmark, and we do not
draw quantitative conclusions from it.

\subsection{Ablation and Mechanistic Analysis}

Table~\ref{tab:Ab_results} evaluates individual components on Go1 under a
$1.15\times$ mass increase. Directional alignment and transient-sensitive
filtering dominate recovery: removing directional alignment increases TTR-50
from 168 to 3367 steps, since the residual can then oppose nominal stabilizing commands, and removing transient-sensitive filtering raises it to 1127. The
auxiliary update mechanisms provide smaller gains, with the dual-timescale
decomposition ($168\rightarrow186$), persistence gate ($168\rightarrow174$),
and climbing-fiber term ($168\rightarrow171$) each changing TTR-50 modestly
and olivary feedback ($168\rightarrow273$) somewhat more. The controls that
regulate where and when corrective authority is applied therefore matter more
than the ones that add representational or update-rule structure. The
factorial analysis in Section~\ref{sec:factorial} provides the direct
comparison of corrective authority against residual capacity; the ablation
here removes components from the full model and does not by itself vary
capacity.

\vspace{2pt}
\noindent\textbf{Hyperparameter sensitivity.}
Because the gate is tuned per platform, we test robustness to imperfect
settings by varying four representative parameters, $\kappa$,
$\gamma_{\max}$, $k_\gamma$, and $\eta_f$, by $\pm50\%$ around their Go1
values under the $2.1\times$ friction increase. Supplementary Table~F1
reports TTR-50; all settings remain crash-free, with only minor changes in
AUC and SSR. Together with Fig.~\ref{fig:recovery_trace}(c), these results
show graceful degradation under suboptimal calibration rather than loss of
feasibility.

\section{Discussion and Conclusion}
We cast rapid inference-time recovery as a constrained residual learning
problem in which the governing question is how much corrective authority an
online correction should be granted around a frozen competent policy, not how
expressive it is. A bounded, authority-regulated residual accelerates recovery
without policy updates, system identification, or shift information; a short
analysis shows the directional gate provably reduces interference with the
nominal action, identifying why it is the dominant control; and the Go1 factorial experiment separates the effect of
corrective authority from residual capacity, while the broader
evaluation spans four morphologies and two physical robots. The main limitations are
that corrective authority is bounded by the available actuation and the gate
is calibrated per platform, the directional test is a coordinate-dependent
sign heuristic that presumes the nominal action stays stabilizing, the analysis establishes bounded residual intervention but
not weight convergence or closed-loop ISS, and severe compound shifts, sparse-feedback settings, and
multi-trial mobile-robot hardware validation remain
challenging. A coupling-aware or value-aligned gate and reduced per-platform
calibration are natural next steps.

\bibliography{references}

\end{document}